\documentclass[conference]{IEEEtran}
\IEEEoverridecommandlockouts
\usepackage{cite}
\usepackage{amsmath,amssymb,amsfonts,amsthm}
\usepackage{algorithm2e}
\usepackage{algpseudocode}
\usepackage{graphicx}
\usepackage{textcomp}
\usepackage{multirow}
\usepackage{hyperref}
\usepackage{comment}
\usepackage{booktabs}
\usepackage[table, svgnames, dvipsnames]{xcolor}
\usepackage{tcolorbox}
\usepackage[table, svgnames, dvipsnames]{xcolor}
\def\BibTeX{{\rm B\kern-.05em{\sc i\kern-.025em b}\kern-.08em
    T\kern-.1667em\lower.7ex\hbox{E}\kern-.125emX}}

\newtheorem{lem}{Lemma}

\newtheorem{defn}{\textbf{Definition}}%[section]

\newtheorem{rem}{Remark}
\newtheorem{obs}{Observation}

\makeatletter

\def\ps@IEEEtitlepagestyle{
    \def\@oddfoot{\mycopyrightnotice}
    \def\@evenfoot{}
}
\def\mycopyrightnotice{
}



\makeatletter
\newcommand*\titleheader[1]{\gdef\@titleheader{#1}}
\AtBeginDocument{%
  \let\st@red@title\@title%
  \def\@title{%
    \bgroup\normalfont\small\raggedright\@titleheader\par\egroup
    \vskip0.1em\st@red@title}
} \makeatother

\title{Scaled and Inter-token Relation Enhanced Transformer for Sample-restricted Residential NILM
}

\titleheader{}

\begin{document}

\author{\IEEEauthorblockN{Minhajur Rahman}
\IEEEauthorblockA{\textit{Dept. of Electrical \& Electronic Engineering} \\
\textit{Int'l Islamic University Chittagong}\\
Chittagong, Bangladesh \\
fahad061299@gmail.com}
\and
\IEEEauthorblockN{Yasir Arafat}
\IEEEauthorblockA{\textit{Dept. of Electrical \& Electronic Engineering} \\
\textit{Int'l Islamic University Chittagong}\\
Chittagong, Bangladesh \\
yaeeeiiuc@gmail.com}
}

\maketitle

\begin{abstract}
Transformers have demonstrated exceptional performance across various domains due to their self-attention mechanism, which captures complex relationships in data. However, training on smaller datasets poses challenges, as standard attention mechanisms can over-smooth attention scores and overly prioritize intra-token relationships, reducing the capture of meaningful inter-token dependencies critical for tasks like Non-Intrusive Load Monitoring (NILM). To address this, we propose a novel transformer architecture with two key innovations: inter-token relation enhancement and dynamic temperature tuning. The inter-token relation enhancement mechanism removes diagonal entries in the similarity matrix to improve attention focus on inter-token relations. The dynamic temperature tuning mechanism, a learnable parameter, adapts attention sharpness during training, preventing over-smoothing and enhancing sensitivity to token relationships. We validate our method on the REDD dataset and show that it outperforms the original transformer and state-of-the-art models by 10-15\% in F1 score across various appliance types, demonstrating its efficacy for training on smaller datasets.
\end{abstract}

%\cite{hassan2022differentially} for dynamic power pricing
%\cite{rashid2019evaluation} for anomaly detection 
\section{Introduction}
Energy efficiency is a growing challenge due to increasing energy demands, making effective power management critical \cite{alahakoon2015smart}. Smart grids and smart meters provide real-time data on energy usage, enabling utilities and consumers to implement advanced demand-side management strategies \cite{ali2022demand}. NILM is crucial in home energy management systems (HEMS), disaggregating household energy usage into individual appliance data. This granular insight allows consumers to optimize their energy consumption, reduce wastage, and shift loads to reduce peak demands. Additionally, NILM aids in load forecasting and identifying appliance malfunctions, contributing to an effective, efficient and automated HEMS \cite{dinesh2019residential}.

NILM is framed as a single-channel blind source separation problem, traditionally NILM has been approached using statistical models such as hidden Markov models (HMMs) and conditional random fields (CRFs) \cite{zia2011hidden}. These models are capable of identifying appliance states but face significant challenges in scalability and in generalizing to unseen devices. However, recent advancements in deep learning have significantly enhanced NILM by introducing sequence-to-sequence models, which offer improved accuracy and efficiency. State-of-the-art architectures, including convolutional neural networks (CNNs), recurrent neural networks (RNNs), long short-term memory (LSTM) networks, and transformers, have revolutionized energy management systems by achieving superior performance and facilitating further progress in the field \cite{kelly2015neural, yue2020bert4nilm}.

Transformers have become a cornerstone of modern deep learning due to their ability to model complex relationships across diverse data modalities such as text, images, and signals \cite{vaswani2017attention}. Their self-attention mechanism enables capturing long-range dependencies, making them superior to CNNs in many complex tasks. However, transformers inherently lack local inductive biases like those present in CNNs, relying instead on large-scale datasets to learn effective representations. This limitation presents a significant challenge for domains like NILM, where data collection is often constrained by privacy regulations, hardware limitations, and practical feasibility \cite{adewole2022privacy}. Consequently, there is a pressing need to adapt transformer architectures for optimal performance on small-scale NILM datasets. Existing solutions to this challenge have focused on introducing locality-preserving mechanisms, such as hierarchical architectures or modified self-attention modules, which mimic inductive biases found in structured data. While these approaches offer improvements, they still require medium-sized datasets and significant architectural tuning or expansion, which may not generalize well to NILM’s unique requirements. NILM datasets are characterized by sparsity, overlapping appliance signals, and high variability, all of which exacerbate the challenges faced by standard transformers.

This paper addresses these challenges by designing a transformer architecture tailored for small-scale NILM datasets. Our approach is motivated by the observation that standard attention mechanisms often suffer from \textit{over-smoothing of attention scores} and an \textit{overemphasis on intra-token relationships}, both of which degrade performance when data is limited. To overcome these issues, we aim to improve the granularity of token relationships and the adaptability of attention distributions. The key contributions of this work are: 

\begin{itemize}
    \item \textbf{Inter-token Relation Enhancement:} This mechanism modifies the self-attention computation by removing the influence of diagonal entries in the similarity matrix, thereby effectively reducing intra-token focus and more promoting meaningful interactions between tokens.
    \item \textbf{Dynamic Temperature Tuning:} By introducing a learnable temperature parameter, we enable the model to dynamically and adaptively adjust the sharpness of its attention scores, mitigating over-smoothing and enhancing sensitivity to token relationships during training.
\end{itemize}

These contributions are firmly grounded with mathematical rigor and experimentally validated on the REDD dataset \cite{kolter2011redd}. The proposed mechanisms significantly improve attention mechanism, enabling the transformer to better handle NILM’s sparse and noisy data. Results demonstrate that our model outperforms the original transformer by 10-15\% in F1 score across multiple appliance types, establishing its effectiveness for small-scale NILM datasets.
\subsection{Related Works}
NILM has advanced significantly with deep learning methods, overcoming the limitations of earlier classical machine learning approaches. Earlier widely used models like HMMs and CRFs focused on disaggregating energy use by modeling appliance states \cite{zia2011hidden}. These approaches faced scalability challenges, limited generalization, and computational inefficiencies, making them impractical for real-time use.

%\cite{zhang2018sequence}
Deep learning has proven more effective by learning complex patterns directly from data. CNNs, widely used in NILM, excel at extracting local features from NILM data \cite{kelly2015neural}. Kelly and Knottenbelt demonstrated the advantages of CNNs, LSTMs, and autoencoders, achieving notable performance gains over classical models \cite{kelly2015neural}. Enhanced CNN architectures improved detection of appliance-specific features like power changes and thresholds. However, CNNs struggle to capture long-range dependencies which are critical for accurate disaggregation. Transformers address this limitation with attention mechanisms that model both global and local dependencies \cite{vaswani2017attention}. Yue’s BERT4NILM achieved significant performance gains for NILM using the vanilla transformer \cite{yue2020bert4nilm}. Inspired by this, transformer-based methods have shown effectiveness across three main categories: the first group employs standard architectures with simple training mechanisms \cite{yue2020bert4nilm, yue2022efficient}, \cite{varanasi2024stnilm}, \cite{rahman2024deep}. The second group uses advanced training techniques with simpler architectures \cite{sykiotis2022efficient}, \cite{sykiotis2022electricity}, \cite{xuan2024enhanced}. The third combines advanced training methods with substantial architectural changes \cite{yue2022efficient}, \cite{varanasi2024stnilm}, \cite{he2023infocus}. Transformers require large datasets, which is challenging in NILM due to privacy constraints. Standard architectures also face over-smoothing and excessive focus on intra-token relations, limiting performance on smaller datasets. Our approach addresses these issues by aligning with the first group, proposing a vanilla transformer with improvements tailored for small-scale NILM. An enhanced attention mechanism strengthens inter-token relations and reduces over-smoothing, offering a lightweight yet effective solution. While we do not compete directly with the advanced architectures of the second and third groups, our method is complementary and well-suited for resource-constrained NILM scenarios.
\section{Proposed Method}
\subsection{Problem Formulation}
The original formulation of attention in transformer \cite{vaswani2017attention} is formulated as follows. Suppose, we have a set of tokens $\mathbf{X} \in \mathbb{R}^{n\times d}$, where $d$ is the dimension of token embedding and $n$ is the number of tokens (for simplicity of discussion, we are omitting the description of how these tokens are computed in this section; however, it is discussed in the following section). The attention is computed by applying three projections on $\mathbf{X}$ which produces a query matrix $\mathbf{Q} \in \mathbb{R}^{n \times d_k}$, key matrix $\mathbf{K} \in \mathbb{R}^{n \times d_k}$ and value matrix $\mathbf{V} \in \mathbb{R}^{n \times d_k}$. Formally, these projections are obtained as follows:

\begin{equation}
    \mathbf{Q} = \mathbf{XW}_Q, \quad \mathbf{K} = \mathbf{XW}_K, \quad \mathbf{V} = \mathbf{XW}_V,
\end{equation}
where, where $\mathbf{W}_Q \in \mathbb{R}^{d \times d_k}$, $\mathbf{W}_K \in \mathbb{R}^{d \times d_k}$ and $\mathbf{W}_V \in \mathbb{R}^{d \times d_v}$ are learnable weight matrices without any constraints. The attention is computed via a similarity matrix $\mathbf{S} \in \mathbb{R}^{n \times n}$ between $\mathbf{Q}$ and $\mathbf{K}$, defined as:

\begin{equation}
    \mathbf{S}_{ij} = \mathbf{Q}_i\mathbf{K}_j^\top.
    \label{eq: dot-product}
\end{equation}
$\mathbf{S}_{ij}$ is similarity between the $i$-th query and the $j$-th key, capturing the semantic relationship between these tokens. $\mathbf{S}$ further goes through a normalisation with $\mathtt{softmax}(\cdot)$ operator and scaled by a factor of $\frac{1}{\sqrt{d_k}}$. Formally, the attention $\mathbf{A}$ for the $i$-th token is computed as follows.

\begin{equation}
    \mathbf{A}_i = \sum_{j=1}^{n} \mathtt{softmax}\left(\frac{\mathbf{S}_{ij}}{\sqrt{d_k}}\right) \mathbf{V}_j
    \label{eq: self-attention}
\end{equation}
where $\mathbf{A}_i \in \mathbb{R}^{d_v}$ is the attention matrix for the \( i \)-th token, and the $\mathtt{softmax}(\cdot)$ operator is applied row-wise to the scaled similarity matrix $\mathbf{S}$. Note that $\mathbf{A}$ reflects self-attention scores.

\begin{tcolorbox}[boxrule=1pt, left=1mm, right=1mm, top=1mm, bottom=1mm]
In Eq. (\ref{eq: self-attention}), the scaling factor $\sqrt{d_k}$ plays a key role during transformer training for stabilizing the gradient since the dot product between query and key in Eq. (\ref{eq: dot-product}) tends to increase excessively. A smaller-valued $d_k$ effectively prevents that. However, this scaling can also lead to undesirable effects in practice. Based on the literature, we find that the entries of the attention matrix $\mathbf{A}$ tend to become similar to each other as training progresses, regardless of the underlying token relationships. This phenomenon can undermine the attention mechanism's ability to discriminate between important and unimportant token relationships. 
\end{tcolorbox}
We identify two potential causes for this issue which are given below in detail:

\begin{figure*}[t]
\centerline{\includegraphics[width=1\linewidth]{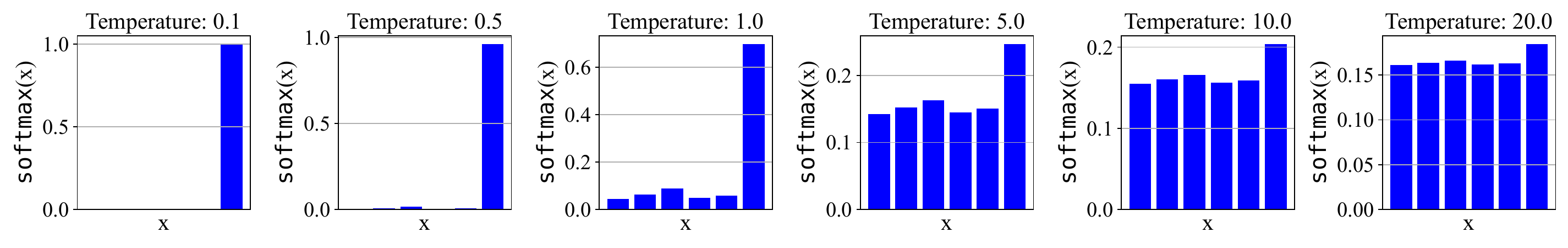}}
\caption{Effect of scaling factor \( \sqrt{d_k} \) on the attention score distribution. Higher values of \( d_k \) lead to smoother distributions, which reduce the model's ability to capture meaningful relationships. Note that x is a randomly sampled array and $x = \mathtt{\{0.1081, 0.4376, 0.7697, 0.1929, 0.3626, 2.8451\}}$.}
\label{fig:Logits}
\end{figure*}

\begin{obs}
\textbf{Projection Redundancy} (\textit{Similarity in Query and Key Projections}). Since both $\mathbf{Q}_i$ and $\mathbf{K}_i$ are obtained via linear projections to the same input $\mathbf{X}_i$, the resulting token embedding naturally exhibit high similarity. Consequently, the similarity matrix $\mathbf{S}_{ij} = \mathbf{Q}_i \mathbf{K}_j^\top$ tends to have larger diagonal entries compared to off-diagonal entries. This can be mathematically expressed as:
\[
\mathbf{S}_{ii} = \mathbf{Q}_i \mathbf{K}_i^\top = (\mathbf{X}_i \mathbf{W}_Q) (\mathbf{X}_i \mathbf{W}_K)^\top, \quad \mathbf{S}_{ij} = \mathbf{Q}_i \mathbf{K}_j^\top,
\]
where $\mathbf{S}_{ii}$ is typically larger than $\mathbf{S}_{ij}$ for \( i \neq j \) due to the shared linear transformation over \( \mathbf{X}_i \). As a result, the $\mathtt{softmax}(\cdot)$ operator disproportionately favors the diagonal entries, i.e., $\mathtt{softmax}(\mathbf{S}_{ii}) \gg \mathtt{softmax}(\mathbf{S}_{ij}) \quad \text{for} \quad i \neq j.$

This results in attention being focused primarily on the individual token itself, rather than on meaningful interactions between tokens, thus resulting in degraded model performance.
\end{obs}

\begin{obs}
\textbf{Temperature Scaling Effect} (\textit{Impact of Scaling by \( \sqrt{d_k} \)}). The scaling factor \( \frac{1}{\sqrt{d_k}} \) in self-attention acts similarly to a temperature parameter in softmax distributions. When the temperature is too high (i.e., \( d_k \) is large), the $\mathtt{softmax}(\cdot)$ operator produces a more uniform distribution. This effect can be quantified by analyzing the derivative of the $\mathtt{softmax}(\cdot)$ operator: 
\begin{equation}
    \frac{\partial \mathtt{softmax}(\mathbf{S}_{ij})}{\partial \mathbf{S}_{ij}} = \mathtt{softmax}(\mathbf{S}_{ij}) \left(1 - \mathtt{softmax}(\mathbf{S}_{ij})\right). 
\end{equation}

As \( \mathbf{S}_{ij} \to 0 \), which occurs when \( \frac{\mathbf{S}_{ij}}{\sqrt{d_k}} \) is heavily scaled, the softmax output becomes nearly uniform:
\begin{equation}
    \lim_{\mathbf{S}_{ij} \to 0} \mathtt{softmax}(\mathbf{S}_{ij}) = \frac{1}{n}.
\end{equation}
This uniformity leads to a loss of distinction between important and unimportant token relationships. Figure \ref{fig:Logits} illustrates this effect with varying values of \( d_k \), showing that larger scaling factors tend to smooth the attention scores, resulting in reduced sensitivity to token relations.
\end{obs}

\begin{figure*}
    \centering
    \includegraphics[width=1\linewidth]{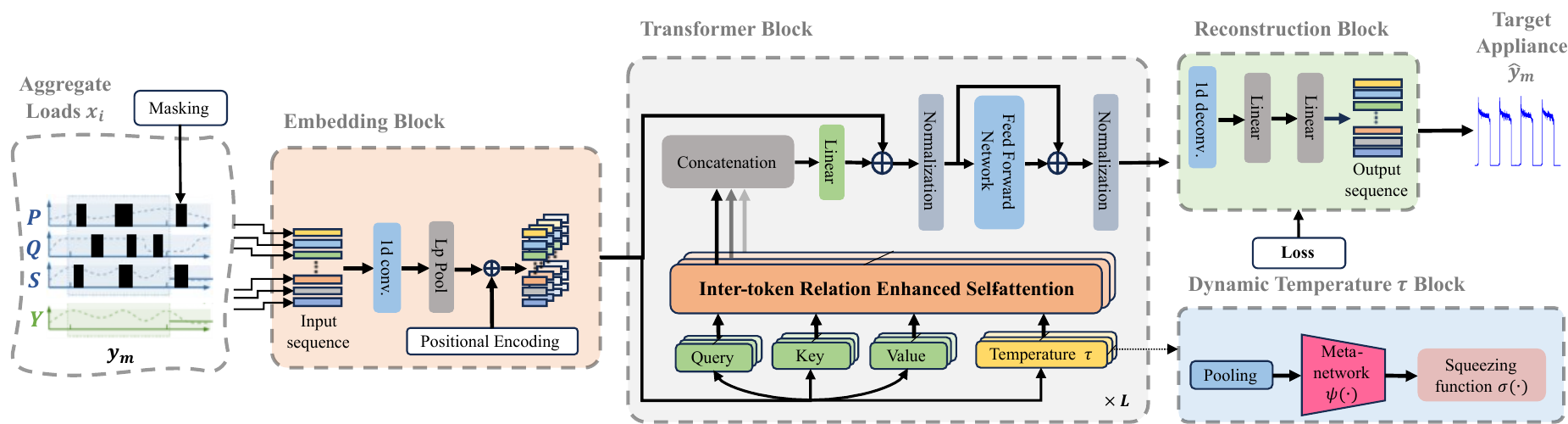}
    \caption{Overview of our proposed method. Given an aggregated load signal, we embed to a higher dimensional space with our embedding block (top left). We then pass it to transformer blocks based on our proposed inter-token relation enhanced self-attention mechanism (middle) and learnable dynamic temperature mechanism (bottom right). Finally, we produce target appliance signals with our reconstruction block (top right). Best viewed in Zoom.}
    \label{fig: main-fig}
\end{figure*}

\subsection{Proposed Solution}

Our method addresses the above two issues with the following solutions. Below we provide a formal and comprehensive explanation of each, underpinned by theoretical foundations.

\subsubsection{Inter-token relation enhancement mechanism}
The self-attention mechanism inherently prioritizes intra-token relationships due to the significant influence of the diagonal entries in the similarity matrix $\mathbf{S}$. This emphasis on self-similarity hinders the model's ability to attend to meaningful inter-token relationships that are crucial for capturing long-range dependencies and contextual information essential for NILM. The proposed inter-token relation enhancement mechanism addresses this limitation by eliminating the diagonal entries of the similarity matrix.

\begin{defn}
   \textbf{\textit{Token similarity matrix}}. Let \( \mathbf{S} \in \mathbb{R}^{n \times n} \) be the token similarity matrix, where each element \( \mathbf{S}_{ij} \) represents the similarity between token \( i \) and token \( j \). The diagonal entries \( \mathbf{S}_{ii} \) represent the intra-token relationships, while the off-diagonal entries \( \mathbf{S}_{ij} \) for \( i \neq j \) represent the inter-token relationships. In standard self-attention, the diagonal entries \( \mathbf{S}_{ii} \), which capture self-similarity, typically dominate the off-diagonal inter-token relations. This dominance leads to an overemphasis on the intra-token attention score.
\end{defn}

\begin{lem}
    \textbf{Impact of Diagonal Entries on Attention Distribution.} Let \( \mathbf{A} \in \mathbb{R}^{n \times n} \) be the attention matrix computed from the softmax-normalized similarity matrix \( \mathbf{S} \). If \( \mathbf{S}_{ii} \) has large positive values, the $\mathtt{softmax}$ operator \( \mathtt{softmax}(\mathbf{S}/\tau) \) will allocate a disproportionately large fraction of attention to the diagonal entries, reducing the model’s capacity to attend to other tokens. Specifically, 

    \begin{equation}
        \lim_{\mathbf{S}_{ii} \to +\infty} \mathbf{A}_{ii} \to 1, \quad \forall i.
    \end{equation}
\end{lem}

To counter this effect, we propose to nullify the diagonal entries, thereby forcing the model to focus on inter-token relationships. We enhance inter-token relationships by replacing the diagonal entries \( \mathbf{S}_{ii} \) with \( -\infty \), effectively removing the intra-token self-similarity from consideration during attention computation: $
\mathbf{S}_{ii} \leftarrow -\infty, \quad \forall \, i = 1, 2, \dots, n.$
This ensures that the attention scores allocated to intra-token relationships approach zero. The resulting attention matrix, \( \mathbf{A} \), is then computed as 
\begin{equation}
    \mathbf{A}_{ij} = \mathtt{softmax}\left( \frac{\mathbf{S}_{ij}}{\tau} \right) \quad \text{for} \, i \neq j, \quad \mathbf{A}_{ii} \approx 0.
\end{equation}

\begin{rem}
    \textbf{Inter-token Focused Attention}. By removing the diagonal entries from the similarity matrix \( \mathbf{S} \), we ensure that the $\mathtt{softmax}$ operator assigns minimal or zero attention to intra-token relationships. As a result, the total attention is distributed entirely across inter-token relationships, i.e., 
    \begin{equation}
        \sum_{j \neq i} \mathbf{A}_{ij} = 1 \quad \text{and} \quad \mathbf{A}_{ii} \approx 0.
    \end{equation}
\end{rem}

This mechanism improves the model's ability to capture and emphasize the relationships between distinct tokens, leading to enhanced modeling of global dependencies across tokens.

\subsubsection{Temperature tuning mechanism}

In the original self-attention formulation, the similarity scores in \( \mathbf{S} \) are scaled by a fixed temperature \( \sqrt{d_k} \), where \( d_k \) is the dimensionality of the key vectors. While this scaling helps prevent excessively large values from dominating the $\mathtt{softmax}$ operator, it lacks flexibility. The optimal temperature may vary across different datasets, appliance types, and training phases, yet the use of a fixed temperature restricts the model’s capacity to adapt to these variations. Therefore, we propose a learnable temperature parameter, \( \tau \), which is dynamically adjusted during training through a meta-network (shown in Fig. \ref{fig: main-fig}).

\begin{defn}
    \textit{\textbf{Temperature-scaled $\mathtt{softmax}$.}} The temperature-scaled softmax function with temperature \( \tau \) applied to the similarity matrix \( \mathbf{S} \) is defined as: $
\mathbf{A}_{ij} = \frac{\exp\left( \frac{\mathbf{S}_{ij}}{\tau} \right)}{\sum_{k=1}^n \exp\left( \frac{\mathbf{S}_{ik}}{\tau} \right)}.$
As \( \tau \to \infty \), the softmax function approaches a uniform distribution, and as \( \tau \to 0 \), the softmax becomes more peaked, concentrating attention on the most significant values in \( \mathbf{S} \).
\end{defn}

\begin{lem}
    \textbf{Effect of Temperature on Attention Distribution.} For any temperature \( \tau > 0 \), the sharpness of the attention distribution is inversely proportional to \( \tau \). Specifically, lower values of \( \tau \) lead to sharper attention, focusing on fewer token relations, while higher values lead to more diffused attention, distributing focus more uniformly across tokens. Thus, the ideal value of \( \tau \) depends on the nature of the task and data.
\end{lem}

To optimize \( \tau \) dynamically, we introduce a meta-network, denoted by \( \psi(\cdot) \), that learns the appropriate temperature based on the input tokens. The meta-network takes the pooled features of the token embeddings \( \mathbf{X} \) at layer $L$ and outputs a value for \( \tau \), which is bounded to prevent extreme values. Specifically, we constrain \( \tau \) within the empirically decided interval \( \left[\frac{\sqrt{d_k}}{8}, 8\sqrt{d_k}\right] \) using a sigmoid-like range squeezing function.

The meta-network architecture consists of several fully connected layers with ReLU activations:
\begin{equation}
\tau = \psi(\mathbf{X}) = \sigma(\mathbf{W}_2 \, \text{ReLU}(\mathbf{W}_1 \, \mathbf{X}_{\text{pooled}})),
\end{equation}
where \( \mathbf{W}_1 \) and \( \mathbf{W}_2 \) are learnable weights, \( \sigma(\cdot)\) denotes the sigmoid activation function defined as $\sigma(x) = \frac{\sqrt{d_k}}{8} \left( \frac{63}{1 + e^{-x}} + 1 \right)$, and \( \mathbf{X}_{\text{pooled}} \) is a average pooled representation of either the token embeddings \( \mathbf{X} \) or the output of previous transformer block \( \mathbf{X}^{L-1} \). The output of the meta-network is then scaled to the desired range.

\begin{rem}
    \textbf{Dynamic Temperature Adaptation.} By learning \( \tau \) via the meta-network, the model can dynamically adjust the sharpness of its attention distributions during training. In the early stages of training, when global context is more important, the meta-network can assign a higher value of \( \tau \), leading to smoother, more distributed attention. In later stages, as the model focuses on more specific token interactions, \( \tau \) can decrease, sharpening the attention focus. 
\end{rem}

This adaptability improves the model's performance across various datasets and tasks, avoiding over-smoothing or overly peaked distributions. Figure \ref{fig: main-fig} shows the framework of our proposed method. Our framework is similar to other transformer based methods such as \cite{yue2020bert4nilm,rahman2024deep} but with the proposed inter-token relation enhancement and dynamic temperature mechanisms. Briefly, we first encode our aggregated load signal with a convolution based encoder module \cite{yue2020bert4nilm}, followed by a number of improved transformer layers with the above two mechanisms. Finally, we produce the appliance signal with a deconvolution based reconstruction module \cite{yue2020bert4nilm}. We used a BERT based training mechanism similar to \cite{yue2020bert4nilm,rahman2024deep}. 

% \usepackage{multirow}
\begin{table*}[]
\caption{Comparison with the existing methods. The top, middle, and bottom parts of the tables show the methods belonging to groups 1, 2, and 3, respectively. All results are quoted from original papers. The boldface indicates the best results. `--' indicates missing results in the respective paper.}
\label{tab: main-table}
\resizebox{\linewidth}{!}{%
\begin{tabular}{lcccccccccccccccc}
\toprule
\multirow{2}{*}{Method} & \multicolumn{4}{c}{\textbf{Refrigerator}} & \multicolumn{4}{c}{\textbf{Washer}} & \multicolumn{4}{c}{\textbf{Microwave}} & \multicolumn{4}{c}{\textbf{Dishwasher}} \\ \cmidrule(r){2-5} \cmidrule(r){6-9} \cmidrule(r){10-13} \cmidrule(r){14-17}
 & Acc & F1 & MRE & MAE & Acc & F1 & MRE & MAE & Acc & F1 & MRE & MAE & Acc & F1 & MRE & MAE \\ \midrule
GRU + \cite{yue2020bert4nilm} & 0.794 & 0.705 & 0.829 & 44.280 & 0.922 & 0.216 & 0.090 & 27.630 & 0.988 & 0.574 & 0.059 & 17.720 & 0.955 & 0.034 & 0.042 & 25.290 \\
LSTM + \cite{yue2020bert4nilm} & 0.789 & 0.709 & 0.841 & 44.820 & 0.989 & 0.125 & 0.020 & 35.730 & 0.989 & 0.604 & 0.058 & 17.390 & 0.956 & 0.421 & 0.056 & 25.250 \\
CNN \cite{yue2020bert4nilm} & 0.796 & 0.689 & 0.822 & 35.690 & 0.970 & 0.274 & 0.042 & 36.120 & 0.986 & 0.378 & 0.060 & 18.590 & 0.953 & 0.298 & 0.053 & 25.290 \\
BERT4NILM \cite{yue2020bert4nilm} & 0.841 & 0.756 & 0.806 & 32.350 & 0.991 & 0.559 & 0.022 & 34.960 & 0.989 & 0.476 & 0.057 & 17.580 & 0.969 & 0.523 & 0.039 & 20.490 \\
RNN \cite{varanasi2024stnilm} & -- & -- & -- & -- & 0.972 & 0.462 & -- & 17.740 & -- & -- & -- & -- & 0.974 & 0.222 & -- & 14.190 \\
Seq2Point \cite{yue2022efficient} & -- & 0.510 & -- & 66.760 & -- & 0.350 & -- & 34.890 & -- & 0.270 & -- & 24.140 & -- & 0.410 & -- & 22.750 \\
Seq2Seq \cite{varanasi2024stnilm} & -- & -- & -- & -- & 0.975 & 0.480 & -- & 23.050 & -- & -- & -- & -- & 0.977 & 0.255 & -- & 11.940 \\
Compact Transformer \cite{rahman2024deep} & 0.848 & 0.765 & 0.830 & 31.160 & 0.988 & 0.633 & 0.021 & 27.470 & 0.987 & 0.496 & 0.049 & 16.300 & 0.961 & 0.542 & 0.038 & 19.030 \\ \midrule

CTA-BERT \cite{xuan2024enhanced} & 0.887 & 0.761 & 0.796 & 30.690 & 0.993 & 0.694 & 0.017 & 18.020 & 0.997 & 0.599 & 0.056 & 17.610 & 0.962 & 0.472 & 0.046 & 22.410 \\
TransformNILM \cite{sykiotis2022efficient} & 0.875 & 0.799 & 0.823 & 32.470 & 0.997 & 0.902 & 0.016 & 23.070 & 0.990 & 0.611 & 0.056 & 16.410 & 0.969 & 0.655 & 0.050 & 24.050 \\
ELECTRIcity \cite{sykiotis2022electricity} & -- & -- & -- & -- & 0.998 & 0.903 & 0.016 & 23.070 & 0.989 & 0.610 & 0.057 & 16.410 & 0.968 & 0.601 & 0.051 & 24.060 \\ \midrule

ELTransformer \cite{yue2022efficient} & -- & 0.510 & -- & 65.520 & -- & 0.400 & -- & 39.350 & -- & 0.290 & -- & 15.820 & -- & 0.440 & -- & 18.180 \\
Switch T/F \cite{varanasi2024stnilm} & -- & -- & -- & -- & 0.981 & 0.486 & -- & 13.990 & -- & -- & -- & -- & 0.993 & 0.274 & -- & 11.540 \\
WindowGRU \cite{varanasi2024stnilm} & -- & -- & -- & -- & 0.973 & 0.475 & -- & 18.050 & -- & -- & -- & -- & 0.978 & 0.258 & -- & 12.000 \\
SGNet \cite{yue2022efficient} & -- & 0.510 & -- & 67.280 & -- & 0.380 & -- & 40.040 & -- & 0.250 & -- & 19.660 & -- & 0.390 & -- & 21.350 \\
LA-InFocus \cite{he2023infocus} & -- & -- & -- & 28.290 & -- & -- & -- & 13.970 & -- & -- & -- & 21.830 & -- & -- & -- & 12.230 \\

 \midrule

\rowcolor{Gainsboro!60}
\textbf{Proposed Method} & 0.884 & \textbf{0.842} & \textbf{0.712} & \textbf{22.140} & 0.996 & 0.712 & 0.018 & 19.930 & 0.995 & \textbf{0.619} & \textbf{0.031} & \textbf{13.930} & 0.979 & \textbf{0.690} & \textbf{0.021} & 14.280 \\ \hline
\end{tabular}}
\end{table*}

\begin{table*}[]
\caption{Ablation study of the proposed method on REDD dataset. The boldface indicates the best results. Best viewed in Zoom.}
\label{tab: ablation-study}

\resizebox{\textwidth}{!}{%
\begin{tabular}{lcccccccccccccccc}
\toprule
\multirow{3}{*}{\textbf{Variation of proposed method}}                                                                       & \multicolumn{4}{c}{\textbf{Refrigerator}} & \multicolumn{4}{c}{\textbf{Washer}}     & \multicolumn{4}{c}{\textbf{Microwave}}  & \multicolumn{4}{c}{\textbf{Dishwasher}} \\ \cmidrule(r){2-5} \cmidrule(r){6-9} \cmidrule(r){10-13} \cmidrule(r){14-17}
& Acc.    & F1     & MRE   & MAE    & Acc.   & F1    & MRE   & MAE    & Acc.   & F1    & MRE   & MAE    & Acc.   & F1    & MRE   & MAE    \\ \midrule

w/o inter-token enhancement and $\tau = \sqrt{d_k}$                                  & 0.848  & 0.765  & 0.830 & 31.160 & 0.988 & 0.633 & 0.021 & 27.470 & 0.987 & 0.496 & 0.049 & 16.300 & 0.961 & 0.542 & 0.038 & 19.030 \\
w/ inter-token enhancement and $\tau = \sqrt{d_k}$                                       & 0.861  & 0.819  & 0.821 & 30.350 & 0.995 & 0.658 & 0.019 & 22.430 & 0.989 & 0.531 & 0.043 & 15.230 & 0.973 & 0.589 & 0.031 & 15.990 \\
w/ inter-token enhancement and $\tau = \sqrt{d_k}/8$ & 0.858  & 0.780  & 0.826 & 31.010 & 0.982 & 0.589 & 0.031 & 40.120 & 0.978 & 0.454 & 0.069 & 19.240 & 0.975 & 0.591 & 0.029 & 15.010 \\
w/ inter-token enhancement and $\tau = \sqrt{d_k}/4$            & 0.867  & 0.839  & 0.759 & 30.220 & 0.998 & 0.684 & 0.020 & 19.010 & 0.989 & 0.524 & 0.041 & 17.230 & 0.978 & 0.642 & 0.025 & 14.000 \\
w/ inter-token enhancement and $\tau = \sqrt{d_k}/2$              & 0.893  & 0.856  & 0.713 & 27.980 & 0.990 & 0.692 & 0.019 & 18.930 & 0.992 & 0.592 & 0.360 & 14.220 & 0.977 & 0.682 & 0.024 & 13.410 \\
w/ inter-token enhancement and  $\tau = 2\sqrt{d_k}$               & 0.862  & 0.810  & 0.769 & 28.930 & 0.990 & 0.677 & 0.026 & 27.830 & 0.989 & 0.541 & 0.040 & 15.020 & 0.965 & 0.550 & 0.030 & 18.810 \\
w/ inter-token enhancement and  $\tau = 4\sqrt{d_k}$               & 0.851  & 0.773  & 0.784 & 29.010 & 0.981 & 0.613 & 0.027 & 38.140 & 0.972 & 0.451 & 0.082 & 20.910 & 0.951 & 0.539 & 0.037 & 22.040 \\
w/ inter-token enhancement and $\tau = 8\sqrt{d_k}$               & 0.822  & 0.679  & 0.832 & 34.820 & 0.980 & 0.568 & 0.049 & 40.340 & 0.970 & 0.341 & 0.109 & 43.010 & 0.950 & 0.493 & 0.057 & 27.000 \\
w/ inter-token enhancement and $\tau$ learned w/o meta-network & 0.801  & 0.598  & 0.892 & 38.380 & 0.972 & 0.512 & 0.063 & 43.290 & 0.968 & 0.449 & 0.059 & 20.130 & 0.967 & 0.513 & 0.049 & 30.000 \\ \midrule

\rowcolor{Gainsboro!60}
w/ inter-token enhancement and $\tau$ learned w/ meta-network & \textbf{0.884}  & \textbf{0.842}  & \textbf{0.712} & \textbf{22.140} & \textbf{0.996} & \textbf{0.712} & \textbf{0.018} & 19.930 & \textbf{0.995} & \textbf{0.619} & \textbf{0.031} & \textbf{13.930} & \textbf{0.979} & \textbf{0.690} & \textbf{0.021} & 14.280 \\
\bottomrule
\end{tabular}}
\end{table*}

\section{Experimental Settings and Results}
\subsection{Dataset and evaluation protocols}
We use the REDD dataset \cite{kolter2011redd}, which includes power consumption data from six houses (i.e., residential) in USA. The dataset contains time-series data for both whole-home and appliance-specific channels, recorded at high (15 kHz) and low (1 Hz) frequencies. For evaluation, we focus on low-frequency recordings and follow standard data processing protocols from previous studies \cite{yue2020bert4nilm,sykiotis2022electricity,kelly2015neural}. To evaluate model generalization, data from houses 2 to 6 is used for training, and house 1 for testing. We select common appliances across all households, excluding devices with limited presence or faulty data. The four appliances used are the fridge, washer, microwave, and dishwasher.

\subsection{Network settings (architecture) and implementation}
We train separate models for each appliance using a transformer network consisting of an embedding block, a transformer encoder of multiple layers, and a reconstruction block. 
The embedding block has a convolution layer (kernel size=5, padding=2), and the reconstruction block uses a deconvolution layer (kernel size=4, stride=2, padding=1). We use similar losses and their combinations as \cite{yue2020bert4nilm}. The transformer network is implemented in PyTorch. Models were trained for 100 epochs with the AdamW optimizer and BERT-style losses \cite{yue2020bert4nilm} on a P100 GPU from a cloud HPC. Following \cite{rahman2024deep}, we use 2 transformer layers/heads, a hidden dimension of 16, dropout ratio of 0.5, and masking ratio of 0.3 for optimal performance.

\subsection{Evaluation metrics}
We used four common metrics for evaluation: \textit{Accuracy} (Acc.), \textit{F1 score}, \textit{Mean Relative Error} (MRE), and \textit{Mean Absolute Error} (MAE) \cite{yue2020bert4nilm,sykiotis2022electricity,kelly2015neural}. \textbf{1) Acc.} Accuracy is the ratio of true positives (TP) and true negatives (TN) over the total number of predictions, $\text{Acc.} = \frac{\text{TP + TN}}{\text{TP + TN + FP + FN}}$, where FP and FN are false positives and false negatives. \textbf{2) F1 score:} F1 evaluates the model’s performance with imbalanced classes, $\text{F1} = \frac{\text{TP}}{\text{TP} + \frac{1}{2}(\text{FP + FN)}}$. \textbf{3) MRE:} MRE measures the accuracy of appliance energy estimates, $\text{MRE} = \frac{1}{\max(Y)} \sum_{i=1}^{N} |\hat{y_i} - y_i|$. \textbf{4) MAE:} MAE calculates the average prediction error, $\text{MAE} = \frac{1}{N} \sum_{i=1}^{N} |\hat{y_i} - y_i|$. $N$ is the size of batch and $\hat{y_i}$ and $y_i$ are original and reconstructed inputs, respectively.

\subsection{Experimental Results}
Table \ref{tab: main-table} shows the experimental results of our proposed method. We compare our results with 16 recent SOTA (state-of-the-art) methods under four evaluation metrics. Specifically, we compare three groups of methods. The first group uses standard deep networks with relatively simpler training mechanisms and the methods in this group are GRU+ \cite{yue2020bert4nilm}, LSTM+ \cite{yue2020bert4nilm}, CNN \cite{yue2020bert4nilm}, BERT4NILM \cite{yue2020bert4nilm}, Seq2Point \cite{yue2022efficient}, Seq2Seq \cite{varanasi2024stnilm}, RNN \cite{varanasi2024stnilm} and Compact Transformer \cite{rahman2024deep}. The second group uses advanced training mechanisms without significant changes in network architecture and the methods in this group are TransformNILM \cite{sykiotis2022efficient}, ELECTRIcity \cite{sykiotis2022electricity} and CTA-BERT \cite{xuan2024enhanced}. The third group uses advanced training mechanisms with significant changes in network architecture and the methods in this group are ELTransformer \cite{yue2022efficient}, Switch T/F \cite{varanasi2024stnilm}, WindowGRU \cite{varanasi2024stnilm}, SGNet \cite{yue2022efficient} and LA-InFocus \cite{he2023infocus}.
It can be seen that our proposed method surpassed the performance of many methods of all three groups across all metrics and appliance types. Several second and third-group methods show strong results indicating the strength of using advanced training and network architectures. In comparison to these methods, our method is trained with a simple training mechanism and still uses a very similar architecture to the original transformer model \cite{yue2020bert4nilm}. Instead of being competitive with them, we think that incorporating our proposed attention improvement mechanisms into their pipeline could further improve their performance since those mechanisms are model and training-agnostic from the point of principle. Specially, we consider our method as more fairly comparable with the original transformer model for NILM \cite{yue2020bert4nilm}.

\begin{table}[]
\caption{Analysis of the computation time. The forward propagation time of transformer modules is given.}
\label{tab: compute-time}
\resizebox{\linewidth}{!}{%
\begin{tabular}{lc}
\toprule
Type & Time (sec.) \\ \midrule
Standard Attention                                                 & 0.081        \\
+ Inter-token enhancement                                 & 0.094       \\
+ Inter-token enhancement and Prefixed $\tau$               & 0.094       \\
+ inter-token enhancement + Learn $\tau$ w/o meta-network & 0.094       \\
+ inter-token enhancement + Learn $\tau$ w/ meta-network  & 0.120 \\
\bottomrule
\end{tabular}}
\end{table}

\subsection{Ablation study}
We conducted an ablation study of our proposed method to assess the effectiveness of inter-token enhancement and dynamic temperature tuning mechanisms. Table \ref{tab: ablation-study} presents the findings across four appliances from the REDD dataset.

\textit{Study on inter-token relation enhancement mechanism:} The results demonstrate that the inter-token relation enhancement mechanism significantly improves performance, outperforming the baseline method and many SOTA approaches.

\textit{Study on dynamic temperature tuning mechanism:} We experimented with several pre-selected $\tau$ values, i.e., $\sqrt{d_k}/8$, $\sqrt{d_k}/4$, $\sqrt{d_k}/2$, $2\sqrt{d_k}$, $4\sqrt{d_k}$, and $8\sqrt{d_k}$, to evaluate the effectiveness of the learning-based dynamic temperature scaling mechanism. The findings indicate a clear impact of temperature scaling across different appliances, with $\sqrt{d_k}/4$ and $\sqrt{d_k}/2$ yielding the best performance, as expected. We also analyzed the learned $\tau$ values, which showed a significant advantage when utilizing our proposed meta-network.

To further validate the effectiveness of our meta-network-based $\tau$ learning mechanism, we also conducted experiment using $\tau$ as a learnable parameter by initializing it randomly and trained in an end-to-end manner, without the meta-network. The results indicate that the meta-network is essential for achieving improved performance, as the alternative approach led to a negative $\tau$ during the learning process, resulting in poorer outcomes.

\subsection{Computational Complexity}
We compare the compute time of our proposed method with the standard attention mechanism \cite{yue2020bert4nilm, rahman2024deep} using a P100 GPU on an HPC cluster. Table \ref{tab: compute-time} shows the comparison. Our method adds a small amount of computation time compared with the standard attention model. However, it can significantly improve the attention module at the expense of this negligible computing cost. Powerful GPUs may improve this condition.
\section{Conclusion and Future Works}
In this paper, we proposed two mechanisms for improving the original transformer model in terms of its capacity. Our experimental results effectively validate that our proposed mechanisms improve the NILM performance on smaller datasets with transformer architecture. We provide theoretical analysis of our proposed mechanisms. In the future, we plan to experiment with larger datasets and optimize compute time.

\bibliographystyle{ieeetr}
\bibliography{references}

\begin{thebibliography}{10}

\bibitem{alahakoon2015smart}
D.~Alahakoon and X.~Yu, ``Smart electricity meter data intelligence for future energy systems: A survey,'' {\em IEEE transactions on industrial informatics}, vol.~12, no.~1, pp.~425--436, 2015.

\bibitem{ali2022demand}
S.~Ali, A.~U. Rehman, Z.~Wadud, I.~Khan, S.~Murawwat, G.~Hafeez, F.~R. Albogamy, S.~Khan, and O.~Samuel, ``Demand response program for efficient demand-side management in smart grid considering renewable energy sources,'' {\em IEEE Access}, vol.~10, pp.~53832--53853, 2022.

\bibitem{dinesh2019residential}
C.~Dinesh, S.~Makonin, and I.~V. Baji{\'c}, ``Residential power forecasting using load identification and graph spectral clustering,'' {\em IEEE Transactions on Circuits and Systems II: Express Briefs}, vol.~66, no.~11, pp.~1900--1904, 2019.

\bibitem{zia2011hidden}
T.~Zia, D.~Bruckner, and A.~Zaidi, ``A hidden markov model based procedure for identifying household electric loads,'' in {\em IECON 2011-37th Annual Conference of the IEEE Industrial Electronics Society}, pp.~3218--3223, IEEE, 2011.

\bibitem{kelly2015neural}
J.~Kelly and W.~Knottenbelt, ``Neural nilm: Deep neural networks applied to energy disaggregation,'' in {\em Proceedings of the 2nd ACM international conference on embedded systems for energy-efficient built environments}, pp.~55--64, 2015.

\bibitem{yue2020bert4nilm}
Z.~Yue, C.~R. Witzig, D.~Jorde, and H.-A. Jacobsen, ``Bert4nilm: A bidirectional transformer model for non-intrusive load monitoring,'' in {\em Proceedings of the 5th International Workshop on Non-Intrusive Load Monitoring}, pp.~89--93, 2020.

\bibitem{vaswani2017attention}
A.~Vaswani, ``Attention is all you need,'' {\em Advances in Neural Information Processing Systems}, 2017.

\bibitem{adewole2022privacy}
K.~S. Adewole and V.~Torra, ``Privacy issues in smart grid data: from energy disaggregation to disclosure risk,'' in {\em International Conference on Database and Expert Systems Applications}, pp.~71--84, Springer, 2022.

\bibitem{kolter2011redd}
J.~Z. Kolter and M.~J. Johnson, ``Redd: A public data set for energy disaggregation research,'' in {\em Workshop on data mining applications in sustainability (SIGKDD), San Diego, CA}, vol.~25, pp.~59--62, Citeseer, 2011.

\bibitem{yue2022efficient}
Z.~Yue, H.~Zeng, Z.~Kou, L.~Shang, and D.~Wang, ``Efficient localness transformer for smart sensor-based energy disaggregation,'' in {\em 2022 18th International Conference on Distributed Computing in Sensor Systems (DCOSS)}, pp.~141--148, IEEE, 2022.

\bibitem{varanasi2024stnilm}
L.~S. Varanasi and S.~P.~K. Karri, ``Stnilm: Switch transformer based non-intrusive load monitoring for short and long duration appliances,'' {\em Sustainable Energy, Grids and Networks}, vol.~37, p.~101246, 2024.

\bibitem{rahman2024deep}
M.~Rahman and Y.~Arafat, ``Towards a deeper understanding of transformer for residential non-intrusive load monitoring,'' 2024.

\bibitem{sykiotis2022efficient}
S.~Sykiotis, M.~Kaselimi, A.~Doulamis, and N.~Doulamis, ``An efficient deep bidirectional transformer model for energy disaggregation,'' in {\em 2022 30th European Signal Processing Conference (EUSIPCO)}, pp.~1536--1540, IEEE, 2022.

\bibitem{sykiotis2022electricity}
S.~Sykiotis, M.~Kaselimi, A.~Doulamis, and N.~Doulamis, ``Electricity: An efficient transformer for non-intrusive load monitoring,'' {\em Sensors}, vol.~22, no.~8, p.~2926, 2022.

\bibitem{xuan2024enhanced}
Y.~Xuan, C.~Pang, H.~Yu, X.~Zeng, and Y.~Chen, ``An enhanced bidirectional transformer model with temporal-aware self-attention for short-term load forecasting,'' {\em IEEE Access}, 2024.

\bibitem{he2023infocus}
J.~He, Z.~Zhang, L.~Ma, Z.~Zhang, M.~Li, B.~Khoussainov, J.~Liu, and L.~Zhu, ``Infocus: Amplifying critical feature influence on non-intrusive load monitoring through self-attention mechanisms,'' {\em IEEE Transactions on Smart Grid}, vol.~14, no.~5, pp.~3828--3840, 2023.

\end{thebibliography}

\end{document}